\documentclass[sigconf]{acmart}
\AtBeginDocument{%
  }

\setcopyright{cc}
\setcctype{by}
\acmDOI{10.1145/3757279.3785550}
\acmYear{2026}
\copyrightyear{2026}
\acmISBN{979-8-4007-2128-1/2026/03}
\acmConference[HRI '26]{Proceedings of the 21st ACM/IEEE International Conference on Human-Robot Interaction}{March 16--19, 2026}{Edinburgh, Scotland, UK}
\acmBooktitle{Proceedings of the 21st ACM/IEEE International Conference on Human-Robot Interaction (HRI '26), March 16--19, 2026, Edinburgh, Scotland, UK}
\received{2025-09-30}
\received[accepted]{2025-12-01}

\usepackage{hyperref}
\usepackage{tikz}
\usepackage{xcolor}
\usepackage{xspace}
\usepackage{multirow}
\usepackage{microtype}
\usepackage{graphicx}
\usepackage{algorithm} 
\usepackage{algpseudocode}

\definecolor{orange}{RGB}{255,127,0}
\definecolor{limegreen}{RGB}{50, 205, 50}
\definecolor{violet}{RGB}{148,0,211}

\newcommand{\oursys}{\texttt{RoboCritics}\xspace}

\newif\ifCOMMENTS
\COMMENTSfalse

\ifCOMMENTS

\else

\fi

\newcommand{\blackcircle}[1]{%
    \begin{tikzpicture}[baseline=(char.base)]
        \node[draw=black, fill=black, circle, inner sep=.6pt, text=white] (char) {\textbf{#1}};
    \end{tikzpicture}%
}

\begin{document}
\title{RoboCritics: Enabling Reliable End-to-End LLM Robot Programming through Expert-Informed Critics}

\author{Callie Y. Kim}
\orcid{0009-0001-4195-8317}
\affiliation{%
  \institution{Department of Computer Sciences University of Wisconsin--Madison}
  \city{Madison}
  \state{Wisconsin}
  \country{USA}
}
\email{cykim6@cs.wisc.edu}

\author{Nathan Thomas White}
\orcid{0009-0000-9414-9647}
\affiliation{%
  \institution{Department of Computer Sciences University of Wisconsin--Madison}
  \city{Madison}
  \state{Wisconsin}
  \country{USA}
}
\email{ntwhite@wisc.edu}

\author{Evan He}
\orcid{0009-0002-6663-8755}
\affiliation{%
  \institution{Department of Computer Sciences University of Wisconsin--Madison}
  \streetaddress{Department of Computer Sciences, University of Wisconsin--Madison}
  \city{Madison}
  \state{Wisconsin}
  \country{USA}
}
\email{ehe6@wisc.edu}

\author{Frederic Sala}
\orcid{0000-0003-0379-2827}
\affiliation{%
  \institution{Department of Computer Sciences University of Wisconsin--Madison}
  \streetaddress{Department of Computer Sciences, University of Wisconsin--Madison}
  \city{Madison}
  \state{Wisconsin}
  \country{USA}
}
\email{fredsala@cs.wisc.edu}

\author{Bilge Mutlu}
\orcid{0000-0002-9456-1495}
\affiliation{%
  \institution{Department of Computer Sciences University of Wisconsin--Madison}
  \streetaddress{Department of Computer Sciences, University of Wisconsin--Madison}
  \city{Madison}
  \state{Wisconsin}
  \country{USA}
}
\email{bilge@cs.wisc.edu}


\begin{abstract}

End-user robot programming grants users the flexibility to re-task robots in situ, yet it remains challenging for novices due to the need for specialized robotics knowledge. 
Large Language Models (LLMs) hold the potential to lower the barrier to robot programming by enabling task specification through natural language. 
However, current LLM-based approaches generate opaque, ``black-box'' code that is difficult to verify or debug, creating tangible safety and reliability risks in physical systems. 
We present \oursys, an approach that augments LLM-based robot programming with expert-informed motion-level critics. 
These critics encode robotics expertise to analyze motion-level execution traces for issues such as joint speed violations, collisions, and unsafe end-effector poses. 
When violations are detected, critics surface transparent feedback and offer one-click fixes that forward structured messages back to the LLM, enabling iterative refinement while keeping users in the loop. 
We instantiated \oursys in a web-based interface connected to a UR3e robot and evaluated it in a between-subjects user study ($n=18$). 
Compared to a baseline LLM interface, \oursys reduced safety violations, improved execution quality, and shaped how participants verified and refined their programs. 
Our findings demonstrate that \oursys enables more reliable and user-centered end-to-end robot programming with LLMs.

\end{abstract}


\begin{teaserfigure}
    \centering
  \includegraphics[width=\textwidth]{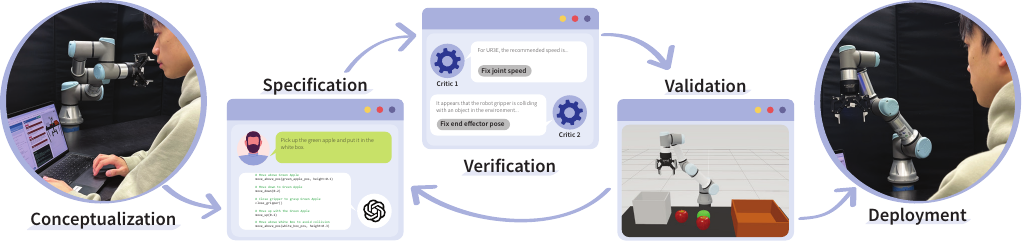}
  \caption{We introduce \oursys, an approach that augments LLM-based robot programming with expert-informed critics, automated fixes, simulation, and physical execution. Critics analyze motion-level execution traces to detect safety and performance issues, surfacing transparent feedback and enabling iterative refinement through one-click, automated fixes while keeping end users in the loop.}
  \Description{The workflow begins with Conceptualization stage then connects to Specification stage by chatting with LLM in natural language. The LLM's generated code then enters the Verification stage, where expert-informed critics analyze the motion-level execution traces. The user then moves to Validation stage via simulation. If not satisfied, the user goes back to Specification stage. If the program is verified, the user is ready for Deployment stage on the physical robot, ensuring the user remains in control throughout the iterative refinement cycle.}
  \label{fig:teaser}
\end{teaserfigure}


\maketitle

\section{Introduction}
Robots are increasingly deployed alongside people in domains such as manufacturing, logistics, healthcare, offices, retail, homes, and transportation \cite{10.1561/2300000066}. Because it is infeasible to pre-program robots for all scenarios, end-user programming enables users to re-task robots in situ, adapting a robot's behavior or goals to meet new task requirements. However, many intended end users (\textit{e.g.,} IT specialists, lab technicians, healthcare professionals) have domain expertise but lack experience with physical robot control, including motion constraints and safety considerations. This challenge is especially important for collaborative robots, where safety verification and interpretability are critical, motivating the need for support mechanisms that help users express intent, interpret robot behavior, and iteratively refine robot behavior.

Large Language Models (LLMs) have increasingly been adopted for robotics applications to translate natural-language task descriptions into executable robot programs, lowering barriers for end-user robot programming. \cite{Kim_2024,10500490}. 
However, deploying LLMs in robotics introduces unique challenges.
Programming robots requires reasoning about physical interactions within complex and dynamic environments, where erroneous instructions carry tangible safety and reliability risks \cite{6630576, GUIOCHET201743, colgate2008safety}. 
LLM generated code is produced through opaque, ``black-box'' reasoning and are often difficult for users, especially novices, to verify, refine, or anticipate unsafe behavior \cite{10.1145/3586030, 10.1145/3589996}.
Verification is further complicated by the reliance of LLMs on predefined API calls that link high-level commands to low-level motor skills \cite{Chen_Huang_2024, 10.1145/3610977.3634969, mower2024rosllmrosframeworkembodied, Song_2023_ICCV}. Known limitations such as hallucination \cite{10.1145/3571730}, compound these challenges, raising concerns about reliability and safety in real-world robotics applications \cite{kambhampati2024llmscantplanhelp, stechly2023gpt4doesntknowits, NEURIPS2023_efb2072a, gou2024criticlargelanguagemodels}. Existing approaches often restrict user involvement to providing input to LLMs and observing output execution \cite{singh2022progpromptgeneratingsituatedrobot, yu2023languagerewardsroboticskill}, leaving critical gaps in verification and debugging.

These outstanding challenges raise the question of how to design mechanisms that allow end users to program robots with LLMs while ensuring safety, transparency, and reliability of their generated programs. While prior work has explored symbolic verification with formal logic or secondary LLMs \cite{10611447, obi2025safeplanleveragingformallogic, khan2025safetyawaretaskplanning}, many safety and performance issues, such as collisions or excessive joint speeds, emerge at the motion level, beyond what code- or symbol-level checks can capture. To address this gap, we investigate the use of \textit{expert-informed critics}: verifiers that encode robotics knowledge and operate directly over execution traces to detect and fix unsafe or inefficient robot behaviors to support end-to-end robot programming, generating and validating executable robot programs directly from conceptualization to deployment. Thus, we pose the following research questions: (1) \textit{How can expert-informed critics be designed to verify LLM-generated robot programs}; (2) \textit{How does integrating expert-informed critics improve the safety and reliability of end-to-end LLM-based robot programming?}

To address these questions, we propose \oursys, an approach that augments LLM-based robot programming with robotics-informed verification and automated fixes. \oursys adapts the Retrieval-Augmented Generation (RAG)-Modulo framework \cite{jain2024ragmodulosolvingsequentialtasks} to incorporate historical task context and introduces external critics that analyze robot motion trajectories for safety and performance issues based on a model of robotics expertise \cite{9889345}.
When violations occur, \oursys surfaces warnings and provides one-click options for users to give LLMs structured feedback for iterative code refinement. By closing the loop between natural language task specification, verification, and execution, \oursys enables end users to maintain both control and confidence in LLM-generated robot behaviors.

We instantiated \oursys in a web-based interface connected to a UR3e robot and evaluated it in a between-subjects-design study. Compared to a baseline LLM interface without critics, \oursys reduced safety violations, improved task execution quality, and shaped how participants verified and refined their programs. In summary, our main contributions include the following:
\begin{enumerate}
    \item \oursys, an approach that integrates LLM-based task specification with expert-informed motion-level critics and automated fixes, enabling users to inspect, understand, and approve corrections during robot execution, with a prototype instantiation\footnote{Code available at \url{https://github.com/Wisc-HCI/RoboCritics}}. 
    \item The design of motion-level critics, which formalize robotics expertise as constraint checks and provide structured feedback to guide verification and refinement. 
    \item Empirical evaluation on a physical robot, demonstrating that \oursys improves reliability, and supports reliable end-to-end robot programming compared to a baseline LLM interface. 
    \item Design implications for integrating expert-informed motion-level critics and automated fixes into LLM-based end-user robot programming systems.
\end{enumerate}

\section{Related Work}
\paragraph*{End-User Robot Programming}
End-user robot programming aims to empower non-experts to adapt and re-task robots without requiring extensive robotics expertise. To lower the barrier to entry, researchers have developed paradigms such as block-based or visual programming \cite{9212036, 10.1145/3610977.3637477, 10.1145/3610978.3640644, 10974063, 10.1145/3610977.3634974}, Programming by Demonstration (PbD) \cite{7745110, doi:10.1177/0278364919884623, 10.1145/3371382.3378300, 10.1145/2701973.2702007}, and natural language interfaces \cite{10.1145/3610978.3640653, 10.1145/3654777.3676401, 7737032, 10.1108/01439910510629244}. 

Despite these advances, robot programming remains uniquely difficult compared to general software development. Programs must account for embodiment-specific challenges such as motion constraints, task objects, obstacles, and interactions with humans \cite{10.1145/2909824.3020215, 8956327, 10.1145/3332165.3347957}. Effective robot programming requires expertise in areas such as kinematics, planning, and control, which are knowledge that cannot reasonably be expected of end users with diverse technical backgrounds \cite{doi:10.1177/2158244020958736}. The central challenge is thus to distill robotics complexity into methods that are both approachable and robust enough for safe real-world deployment. This gap motivates exploration of new paradigms for end-user robot programming.  

\begin{figure*}[t!] 
    \centering
    \includegraphics[width=\textwidth, alt={\oursys workflow: (1) The user begins by providing a high-level task description. (2) The LLM generates a corresponding robot program. (3) The program is executed and evaluated against user-selected critics based on the resulting trajectory. (4) Feedback and results are stored in the interaction memory. (5) The LLM uses RAG to refine the program. (6) The refined program incorporates critic feedback. (7) The user validates the improved program via simulation. (8) Once verified, the final program is stored for future reference. (9) The validated program is deployed to the physical robot.}]{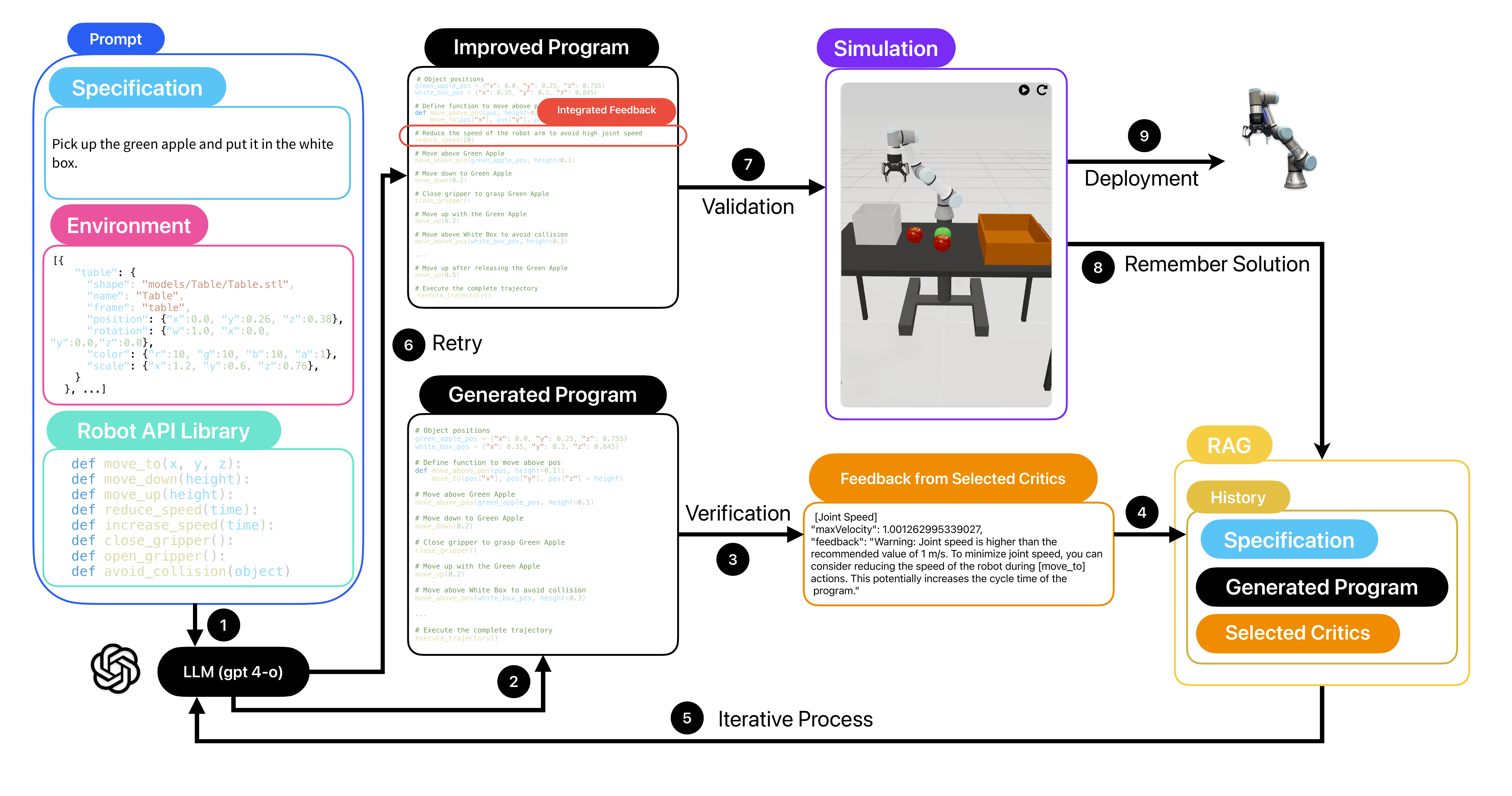}
    \caption{\oursys workflow: (1) The user begins by providing a high-level task description. (2) The LLM generates a corresponding robot program. (3) The program is executed and evaluated against user-selected critics based on the resulting trajectory. (4) Feedback and results are stored in the interaction memory. (5) The LLM uses RAG to refine the program. (6) The refined program incorporates critic feedback. (7) The user validates the improved program via simulation. (8) Once verified, the final program is stored for future reference. (9) The validated program is deployed to the physical robot.}
    \Description{
    \oursys workflow: (1) The user begins by providing a high-level task description. (2) The LLM generates a corresponding robot program. (3) The program is executed and evaluated against user-selected critics based on the resulting trajectory. (4) Feedback and results are stored in the interaction memory. (5) The LLM uses RAG to refine the program. (6) The refined program incorporates critic feedback. (7) The user validates the improved program via simulation. (8) Once verified, the final program is stored for future reference. (9) The validated program is deployed to the physical robot.
    }
    \label{fig:flow}
\end{figure*}

\paragraph*{LLMs for Programming and Robotics}
LLMs have recently emerged as a promising paradigm for lowering barriers to programming by translating natural language into code (NL2Code). Tools such as Github Copilot\footnote{\url{https://github.com/features/copilot}}, Claude\footnote{\url{https://claude.ai}}, and Amazon CodeWhisperer\footnote{\url{https://aws.amazon.com/codewhisperer}} exemplify this trend \cite{jiang2024survey}, and recent evaluations report promising accuracy rates across platforms \cite{athiwaratkun2023multilingualevaluationcodegeneration, yetiştiren2023evaluatingcodequalityaiassisted}. Despite these gains, challenges persist: LLMs frequently hallucinate, lack domain-specific depth, and cannot reliably self-correct \cite{10.1145/3708359.3712091, xu-etal-2024-pride, wataoka2025selfpreferencebiasllmasajudge, kambhampati2024llmscantplanhelp}. Collectively, this body of work suggests that LLMs, when used in isolation, are insufficient for reliable deployment.

Robotics has emerged as a key application area where natural language interfaces promise to lower barriers for non-experts. LLMs have been applied to task planning \cite{ahn2022icanisay, driess2023palme, 10802322}, human-robot collaboration \cite{Kodur_2023, 10141597, 10.1145/3610977.3634966, 10561501, 10.1145/3610977.3634999}, and multimodal interaction \cite{10974232}. Their ability to sustain dialog and maintain context allows iterative task refinement \cite{10500490}, making robot programming more conversational and accessible. Yet, translating free-form language into executable robot code carries significant risks: errors in generated programs manifest physically, and the opaque reasoning of LLMs leaves safety and correctness difficult to assess.  

To address these risks, researchers have explored formal methods such as linear temporal logic (LTL) for correctness \cite{10611447, rabiei2025ltlcodegencodegenerationsyntactically}, as well as frameworks like RAG-Modulo that introduce external verifiers (``critics'') to check syntax and semantics \cite{jain2024ragmodulosolvingsequentialtasks}. However, most of these efforts focus on evaluation benchmarks or planning-level correctness, leaving open questions about how verification mechanisms can be designed for interactive, end-to-end robot programming with end users in the loop. Recent user-centered systems highlight this gap, illustrating the difficulty with ambiguity in natural language input \cite{10974179} and demonstrating the need for program verification \cite{10.1145/3610977.3634969}.
Building on this work, we extend critic-based approaches to the motion level, where issues like collisions or joint speed limits arise, by integrating structured feedback and automated fixes directly into an interactive LLM programming workflow.

\section{Technical Approach}


Our approach is based on the premise that LLMs alone cannot ensure safe or reliable robot programs due to their lack of grounded physical understanding. We introduce a general framework that augments LLM-based programming with expert-informed critics as modular verifiers that analyze executed trajectories and provide structured feedback.
In \oursys, we focus on external, motion-level critics adapted from expert-encoded frames \cite{9889345}, enabling detection and correction of safety-critical errors while remaining extensible to additional constraints and domains.
To support transparency and alignment, we incorporate a human-in-the-loop (HITL) review for critic-proposed fixes. This allows users to inspect and approve corrections, ensuring updated robot behaviors remain consistent with their task intent.


\subsection{Workflow Overview}

Figure \ref{fig:flow} illustrates the system workflow through a representative scenario. The user initiates the process by providing a high-level task description to the LLM, for example, putting a green apple in the white box (Step \blackcircle{1}). The LLM (gpt-4o) generates a corresponding robot program  using environment information and a library of predefined robot action APIs (Step \blackcircle{2}). This program is then executed, during which user-selected critics assess the resulting trajectory for issues such as collision risks, unsafe gripper poses, or excessive joint speed (Step \blackcircle{3}). The verification output, along with the user request and program, is stored in an interaction memory (Step \blackcircle{4}). Using RAG, the LLM accesses this memory to refine the program in response to critic feedback (Step \blackcircle{5}). For instance, if a joint speed warning was issued, the updated program may include a reduce\_speed function (Step \blackcircle{6}). Users can then simulate the refined program to validate whether it aligns with their intent (Step \blackcircle{7}). Once validated, the final version of the program is stored for future reference (Step \blackcircle{8}) and can be deployed to the physical robot (Step \blackcircle{9}). This loop allows users to iteratively interpret, validate, and improve LLM-generated robot code, enhancing performance, safety, and alignment with task goals.

\subsubsection{From Execution Trace to Flags}

When a user runs the generated program (Figure \ref{fig:flow} Step \blackcircle{3}), it produces a trajectory composed of robot states, capturing the robot's configuration at each timestep. Using Lively \cite{10.1145/3568162.3576982}, each state includes joint angles, Cartesian frames of links, pairwise link proximity, and a timestamp. Formally, the robot state at timestep $t$ is represented as $s_t = \{J_t, F_t, P_t, \tau_t\}$, where $J_t$ denotes the joint angles, $F_t$ the link frames, $P_t$ the pairwise link proximity, and $\tau_t$ the timestamp. Each critic $\mathcal{C}_i$ operates on the trajectory, which is a sequence $\{s_1, \dots, s_T\}$ and outputs a flag, \textit{OK}, \textit{Warning}, or \textit{Error}, with a natural language explanation and hints for addressing the concern.


\subsubsection{Expert-Informed Critics}
\oursys includes five critics acting as analytic functions that examine motion traces: The \textbf{space-usage} critic evaluates the convex hull of link positions across the trajectory; if the occupied volume exceeds 50\% of the workspace $\mathcal{W}_{\text{allowed}}$, it returns \textit{Warning}, and if it exceeds the boundary, \textit{Error}. The \textbf{collision} critic evaluates the proximity between the gripper geometry and environment objects using an axis-aligned bounding box (AABB) distance check. It returns \textit{Error} on penetration, \textit{Warning} when the distance is less than threshold $d_{\text{warn}}$, and \textit{OK} otherwise. The \textbf{joint speed} critic estimates joint speed $v_j$ using the Cartesian (linear) speed of a reference point on the corresponding link in the world frame, as a proxy for angular joint velocity. It issues a \textit{Warning} if $v_j > v_{\text{warn}}$, and an \textit{Error} if $v_j > v_{\text{max}}$, where $v_{\text{warn}}$ is the recommended safe threshold and $v_{\text{max}}$ is the robot's maximum allowable speed (see Algorithm~\ref{alg:jointspeed}). The \textbf{end-effector pose} critic quantifies the risk of ``spearing'' (the gripper moves quickly in the direction of its fingers) by computing a pose score $ps_i$ for each step $i$. This score is derived from the angle $\theta_i$ between the gripper's motion vector $\Delta \mathbf{p}_i$ and its finger direction $\mathbf{d}_i$, alongside the motion speed. Rapid, highly aligned motions yield high scores. Thus it issues a \textit{Warning} if the maximum score $ps_{\max} \geq \text{score}_{\text{warn}}$ or an \textit{Error} if $ps_{\max} \geq \text{score}_{\text{err}}$. The \textbf{pinch-point} critic monitors the pairwise link proximity ($P_t$) derived from the robot state $s_t$. If proximity value $d < d_{\min}$, it issues an \textit{Error}, indicating a dangerous pinch region. If $d_{\min} \leq d < d_{\max}$, it returns a \textit{Warning}, while $d \geq d_{\max}$ is considered \textit{OK}.

\begin{algorithm}[t]
\caption{Example critic: \textsc{Joint Speed Critic}.}
\label{alg:jointspeed}
\begin{algorithmic}[1]
\Procedure{Evaluate}{$\{J_t\}_{t=1}^{T}$, $v_{\text{warn}}$, $v_{\text{max}}$}
\Comment{$\{J_t\}_{t=1}^{T}$ denotes the sequence of joint angles from the trajectory}
\State flag $\gets$ \textit{OK}
\For{$t \gets 2$ \textbf{to} $T$}
    \State $\Delta t \gets (\tau_t - \tau_{t-1}) / 1000$ 
    \State $F_t \gets \textsc{Frames}(J_t)$
    \ForAll{$l \in \mathrm{keys}(F_t)$}
        \State $p_{t-1} \gets \textsc{WorldPos}(F_{t-1}^l)$
        \State $p_t \gets \textsc{WorldPos}(F_t^l)$
        \State $v_j \gets \lVert p_t - p_{t-1} \rVert / \Delta t$
        \If{$v_j > v_{\text{max}}$}
            \State \Return \textit{Error}
        \ElsIf{$v_j > v_{\text{warn}}$}
            \State flag $\gets$ \textit{Warning}
        \EndIf
    \EndFor
\EndFor
\State \Return flag \Comment{\textit{OK} if no threshold is exceeded}
\EndProcedure
\end{algorithmic}
\end{algorithm}

\subsubsection{Extensible and modular} 
A key strength of the \oursys architecture is its extensibility and modularity. The implementation of critics as independent, modular verifiers that operate solely on execution traces decouples the safety verification layer from the LLM's program generation step. This technical distinction is fundamental: it allows new safety and performance constraints, formalized as critics, to be introduced or updated without altering the underlying language model or the robot's core skill library. 
Critics are designed to be verifiable against robotics domain constraints and to generate actionable feedback that supports program revision. 
This design enables accommodation of domain- and abstraction-specific critics (\textit{e.g.,} force limits in industrial settings or routine adherence in healthcare), and supports continuous adaptation to evolving tasks and safety requirements.

\subsubsection{Automated Fixes} 
\begin{figure*}[t] 
    \centering
    \includegraphics[width=\textwidth, alt={User interface of \oursys: The panel on the left hosts a chat interface (a) for natural language interaction with the LLM agent. Users can chat by sending a message (i). The center panel exposes critics (b) allowing users to inspect and verify the generated program before execution. Users can activate critics by selecting them (g) and also fix the issues raised by critics by pressing the fix button (f). After selecting critics, users can execute the generated code by pressing ``Run Code'' button (d). The terminal shows the feedback from the selected critics (h). The panel on the right displays a simulation (c) of the generated robot program, allowing users to play and replay the execution as needed (e).}]{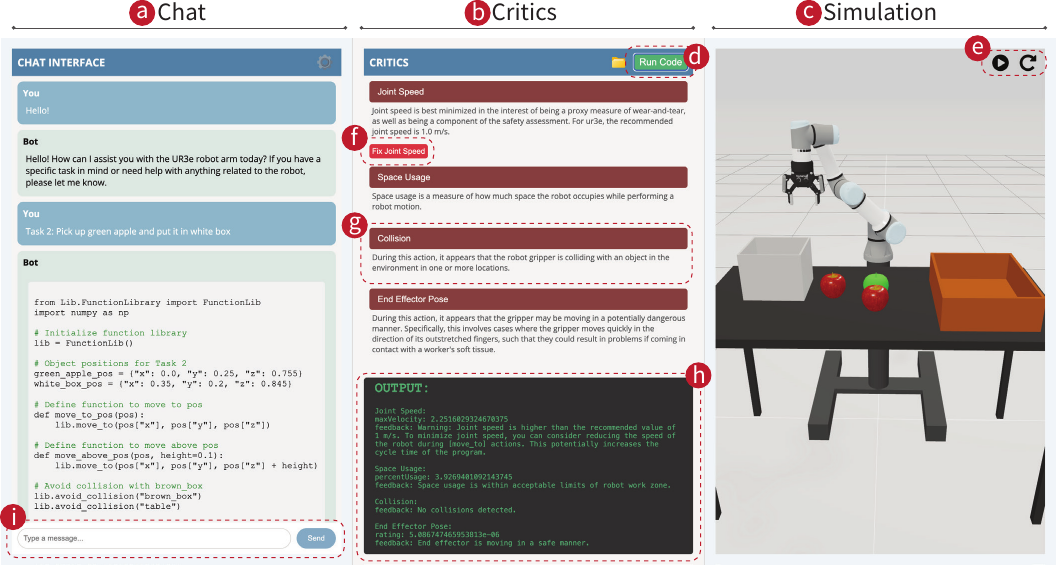}
    \caption{User interface of \oursys: The panel on the left hosts a chat interface (a) for natural language interaction with the LLM agent. Users can chat by sending a message (i). The center panel exposes critics (b) allowing users to inspect and verify the generated program before execution. Users can activate critics by selecting them (g) and also fix the issues raised by critics by pressing the fix button (f). After selecting critics, users can execute the generated code by pressing ``Run Code'' button (d). The terminal shows the feedback from the selected critics (h). The panel on the right displays a simulation (c) of the generated robot program, allowing users to play and replay the execution as needed (e).}
    \Description{User interface of \oursys: The panel on the left hosts a chat interface (a) for natural language interaction with the LLM agent. Users can chat by sending a message (i). The center panel exposes critics (b) allowing users to inspect and verify the generated program before execution. Users can activate critics by selecting them (g) and also fix the issues raised by critics by pressing the fix button (f). After selecting critics, users can execute the generated code by pressing ``Run Code'' button (d). The terminal shows the feedback from the selected critics (h). The panel on the right displays a simulation (c) of the generated robot program, allowing users to play and replay the execution as needed (e).}
    \label{fig:interface}
\end{figure*}

When an execution trace is flagged by a critic, the violation is surfaced in the terminal (Figure~\ref{fig:interface}h) together with a structured feedback message describing the issue and a candidate revision (\textit{e.g.}, \textit{``Warning: Joint speed is higher than the recommended value of 1 m/s. To minimize joint speed, you can consider reducing the speed of the robot during [move\_to] actions. This potentially increases the cycle time of the program.''}). If the user presses the Fix button (Figure~\ref{fig:interface}f), this structured message is forwarded to the LLM, which produces a revised program (Figure \ref{fig:flow} Step \blackcircle{6}). The updated code is then displayed in the chat interface (Figure~\ref{fig:interface}a) for inspection and optional re-execution. This design maintains user oversight while reducing manual editing effort. By exposing both the critic's rationale and the automated patch, the interface supports an interpretable verification-fix loop: critics detect violations from motion-level traces, and the LLM translates their guidance into executable revisions. This loop ensures that fixes are grounded in actual robot behavior rather than relying solely on abstract program representations, addressing safety-critical errors that prompt-only verification may miss.

\subsubsection{Integration with RAG}
\oursys utilizes RAG, maintaining historical context across interactions, allowing the LLM to refine code iteratively rather than generating entirely new solutions at each turn (Figure \ref{fig:flow} Step \blackcircle{5}). Specifically, the system stores the user's request, the generated program, and the feedback produced by critics as tuples, ensuring consistent context reuse.

\subsection{Implementation} \label{sec:implementation}
\oursys is instantiated as a web-based interface that integrates LLM-based program generation, motion-level critics, and interactive debugging in a one workflow (Figure~\ref{fig:interface}).

The backend employs \texttt{gpt-4o} \cite{openai2024helloGPT4o} for program generation and \texttt{text-embedding-ada-002} \cite{openai2024embeddingmodel} for RAG using LangChain\footnote{\url{https://github.com/langchain-ai/langchain}}. 
Programs are expressed using a predefined API of high-level commands, such as \texttt{move\_to(x,y,z)},
 \texttt{open\_gripper()}, 
 \texttt{reduce\_speed(time)}, 
and \texttt{avoid\_collision(object)}. These abstractions wrap low-level instructions, allowing the LLM to compose complex behaviors while remaining compatible with the target robot. The backend is built using Python (Flask) to manage program generation, verification, and robot communication.

The system's frontend is implemented in React and renders simulations using Three.js. The task environment is initialized through from a JSON file that specifies the objects' positions, scales, orientations, and identifiers. This structured representation is used to render the simulation scene and support critics in detecting spatial and safety-related violations during execution. 
Critics operate on execution traces derived from the robot's URDF model, generated using the Lively inverse kinematics (IK) solver \cite{10.1145/3568162.3576982}, which provides a Python interface for computing feasible trajectories for the UR3e robotic arm\footnote{\url{https://www.universal-robots.com/products/ur3e/}}. 
For physical execution on the UR3e, verified joint states are transmitted via a dedicated TCP/IP socket connection. This robust setup ensures a smooth transition from the verified simulation to the real-world operation.

\subsection{Validation of the Critics-based Approach}

\begin{table}[t]
\caption{Comparison of embedded (prompt-only) versus external (motion-level) critics across three tasks. The \textit{Attempt} indicates the number of LLM program generation attempts until completion (max 5). The \textit{Score} reports program quality on a $0$-$10$ scale, where higher scores indicate better program quality.}
\Description{A table comparing embedded (prompt-only) and external (motion-level) critics across three tasks recycling, sorting, and preparing breakfast. For each task, the table reports the number of LLM program generation attempts required to complete the task (up to five attempts) and a program quality score on a scale from 0 to 10, where higher scores indicate better quality. External critics consistently score higher than embedded across both number of attempts and score for all three tasks. The average of external critics is higher than embedded for both number of attempts and score.}
\label{tab:embedded-vs-external}
\centering
\begin{tabular}{lcc|cc}
\toprule
 & \multicolumn{2}{c|}{Embedded} & \multicolumn{2}{c}{External} \\
\cmidrule(lr){2-3} \cmidrule(lr){4-5}
 & Attempt & Score & Attempt & Score \\
\midrule
Recycling           & 2 & 5 & 5 & 8 \\
Sorting             & 3 & 8 & 5 & 8 \\
Preparing Breakfast & 2 & 6 & 5 & 7 \\
\midrule
\textbf{Average}       & 2.3 & 6.3 & 5.0 & 7.7 \\
\bottomrule
\end{tabular}
\end{table}

We performed ablations to evaluate the design choice of implementing critics as external, motion-level verifiers rather than embedding their descriptions directly into the LLM prompt. In the \textit{embedded} condition, critic rules were injected into the system prompt with the instruction to review and revise code if any violations were detected (e.g., \textit{``Task: Review the program. If any part might violate these rules, revise the code to fix it.''}). In the \textit{external} condition, critics operated after program generation, analyzing execution traces and surfacing violations with structured feedback for iterative fixes. Both conditions used the same tasks, APIs, and underlying LLM. Prompt details for the embedded condition are provided in the supplementary materials.\footnote{\url{https://osf.io/vdnw7/?view_only=3525789a1462443a884f9c895272ac11}}

Each task (recycling, sorting, preparing breakfast) was tested by repeatedly generating programs up to five attempts, following the rule: if the LLM self-reported that a program was complete and no further code changes were made, attempts were terminated early. Programs were executed in simulation and scored on a $0$-$10$ index derived from expert-informed critics evaluating safety (\textit{e.g.}, collisions, joint speeds, pinch points) and performance (\textit{e.g.}, space usage, end-effector pose). In addition, an experimenter qualitatively examined the generated code to assess whether the embedded LLM demonstrated reasoning consistent with critic rules.  

Table~\ref{tab:embedded-vs-external} reports the results. Embedded critics often converged in fewer attempts but produced lower-quality programs, with violations that went undetected. Embedded critics often ``believed'' their programs were safe despite violations such as collisions, high joint speeds, or unsafe end-effector orientations, which were issues that only external critics detected through execution traces. External critics, by contrast, consistently required the full five iterations but achieved higher program quality scores across all tasks. Qualitative inspection revealed that embedded LLMs frequently described rule compliance in text (\textit{e.g.}, \textit{``Speed was reduced using \texttt{lib.reduce\_speed(25)} to enhance safety.''}) but still failed to produce safe motions. This underscores the limitation of prompt-only verification and highlights how external critics, grounded in motion-level execution traces, enable detection and correction of safety-critical errors. The findings also suggest that providing richer, more structured feedback to the LLM beyond high-level prompt rules can lead to more reliable program revisions.

However, both conditions struggled to resolve pinch point issues, where the robot's motion created unsafe regions that could trap a human hand. This highlights a case where even external critics, despite providing motion-level feedback, could not guarantee correction. This is partially due to the limited number of attempts, the constrained robot API library, and the starting robot configuration. 
\section{User Study}



\paragraph{Study Design}
The study employed a between-subjects design with two conditions: (1) \textit{no-critic}, where participants received no feedback from domain-specific critics, and (2) \textit{with-critic}, where participants received domain-specific feedback from critics and could refine their program using automated fix buttons. All participants interacted with the same implementation of \oursys, described in Section~\ref{sec:implementation}. However, those in the \textit{no-critic} condition did not have access to the Critics panel (Figure~\ref{fig:interface}b). Within each condition, the tasks were assigned in random order.

\paragraph{Tasks}
Participants completed three sequential tasks of increasing complexity: \textit{Recycling}, \textit{Sorting}, and \textit{Preparing Breakfast}. Recycling (easy) involved discarding multiple objects while avoiding collisions. Sorting (medium), framed as apple jam preparation, required attribute-based sorting by ripeness. Preparing Breakfast (hard) chained multiple subgoals and required fine-grained control, state tracking, and debugging of LLM-generated programs under higher task complexity.

\begin{figure}[t]
    \centering
    \includegraphics[width=\linewidth, alt={Initial task state for each tasks performed in the user study. From left to right: recycling, sorting, and preparing breakfast. Recycling task has a coffee cup, container, and a water bottle on the table with a trash bin behind the table. For the sorting task, there are two red apples and one green apple in the middle of the table, with white box on the left and brown box on the right side of the table. For preparing breakfast task, there is a plate with a fork in the middle of the table. On the right, there is a cupcake on top of an apple inside a brown box. A trash bin is behind the table.}]{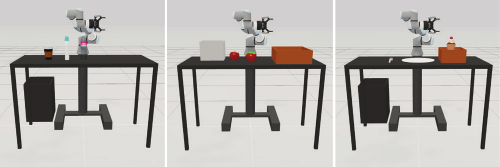}
    \caption{Initial task state for each tasks performed in the user study. From left to right: recycling, sorting, and preparing breakfast.}
    \Description{
    Initial task state for each tasks performed in the user study. From left to right: recycling, sorting, and preparing breakfast. Recycling task has a coffee cup, container, and a water bottle on the table with a trash bin behind the table. For the sorting task, there are two red apples and one green apple in the middle of the table, with white box on the left and brown box on the right side of the table. For preparing breakfast task, there is a plate with a fork in the middle of the table. On the right, there is a cupcake on top of an apple inside a brown box. A trash bin is behind the table.
    }
    \label{fig:task}
\end{figure}

\paragraph{Procedure}
After obtaining informed consent, the experimenter introduced the purpose of the study and outlined the planned activities.
This was followed by a 10-minute tutorial where participants were introduced to \oursys and engaged in a brief hands-on walkthrough. Participants were then assigned to one of two conditions, \textit{no-critic} or \textit{with-critic}, and completed three tasks each with a 10 minute time limit. Upon completing all tasks, participants filled out a post-survey evaluating system usability, cognitive workload, and overall experience. The session concluded with a semi-structured interview asking how participants interacted with \oursys during the tasks. Each study session lasted approximately 60 minutes, and participants were compensated \$$12.50$ per hour. Questionnaires used during the study can be found in the supplementary materials.

\paragraph{Participants}
We recruited 18 participants (9 male, 9 female) through a university mailing list between the ages of 19 and 68 ($M = 39.28$, $STD = 16.34$). Participants were required to be in the United States, fluent in English, and at least 18 years old. All participants agreed to participate in our study via
our institution's IRB-approved consent form.

\paragraph{Measures \& Analysis}
We combined quantitative metrics with qualitative inquiry to assess performance, interaction behavior, and user perception.
To evaluate program quality, each participant-generated program was assessed by five expert-informed critics: joint speed, space usage, end effector pose, collision, and pinch points. Each critic returned a score of 0 (error), 1 (warning), or 2 (no issue), which were summed into a 0-10 program quality index. We used Student's t-tests to compare program scores between the \textit{no‑critic} and \textit{with-critic} conditions. Interaction logs were analyzed to characterize user behavior. 
Subjective measures included NASA-TLX \cite{hart1988development}, System Usability Scale (SUS) \cite{brooke1996sus}, and USE questionnaires \cite{lund2001measuring}.
The USE subscales demonstrated high reliability, with Cronbach's $\alpha$ scores of $0.9522$ (usefulness), $0.9074$ (ease of use), $0.8626$ (learning), and $0.9402$ (satisfaction). 
Student's t-tests and Mann-Whitney U tests were applied based on normality with Shapiro-Wilk tests. Finally, thematic analysis (TA) \cite{McDonald19, clarke2014thematic} of interview data contextualized the quantitative findings.

\section{Findings}
\begin{figure}[t]
    \centering
    \includegraphics[width=\linewidth, alt={There are two subfigures. The left subfigure shows a bar chart with program score from 0 to 10 on the Y axis across three tasks. For each task, there are two conditions, \textit{with-critic} (dark blue) and \textit{no-critic} (light blue). For Task 1 and Task 2, \textit{with-critic} is higher than \textit{no-critic} with significant difference $p < .05$. The right subfigure shows a bar chart with critic counts from 0 to 30 across different critics. In order of descending critic counts is collision, joint speed, space usage, end effector pose, and pinch point.}]{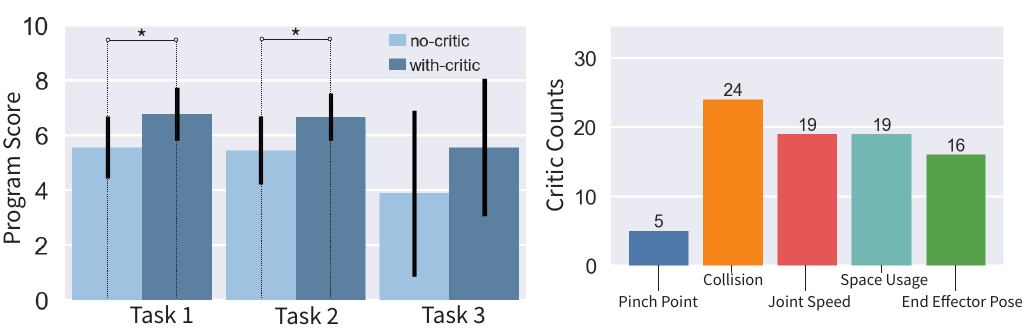}
    \caption{(left) Program quality scores across tasks for both \textit{with-critic} (dark blue) and \textit{no-critic} (light blue) conditions. Scores were computed using domain-specific critics, with higher scores indicating better program quality. Horizontal lines indicate statistically significant differences based on the Student's t-test ($p < .05^{\ast}$). Vertical lines in each bar graph indicate standard deviation. (right) Total number of critic activations across categories.}
    \Description{
    There are two subfigures. The left subfigure shows a bar chart with program score from 0 to 10 on the Y axis across three tasks. For each task, there are two conditions, \textit{with-critic} (dark blue) and \textit{no-critic} (light blue). For Task 1 and Task 2, \textit{with-critic} is higher than \textit{no-critic} with significant difference $p < .05$. The right subfigure shows a bar chart with critic counts from 0 to 30 across different critics. In order of descending critic counts is collision, joint speed, space usage, end effector pose, and pinch point.
    }
    \label{fig:combined_results}
\end{figure}

Having critics had a positive impact on program quality, with participants in critic-assisted conditions consistently achieving higher performance scores (Figure~\ref{fig:combined_results}(left)). In Task 1, critic-assisted participants produced significantly better programs ($M = 6.78$, $SD = 0.97$) compared to those without critics ($M = 5.56$, $SD = 1.13$), $t(16) = 2.46$, $p = .026$, 95\% $CI=[0.17, 2.28]$, $d=1.16$. A similar effect was observed in Task 2, where critic-assisted participants again outperformed those in the non-assisted condition ($M = 6.67$, $SD = 0.87$ vs. $M = 5.44$, $SD = 1.24$), $t(16) = 2.43$, $p = .027$, 95\% $CI=[0.16, 2.29]$, $d=1.15$. In Task 3, although critic-assisted participants showed higher average scores ($M = 5.56$, $SD = 2.51$) than their counterparts ($M = 3.89$, $SD = 3.02$), this difference was not statistically significant, $t(16) = 1.27$, $p = .221$, 95\% $CI=[-1.10, 4.44]$, $d=0.60$. 
The gains in objective program quality were achieved without significantly impacting participants' subjective experience, showing no significant changes in USE, SUS, or NASA-TLX (see supplementary materials).


\paragraph{Perceived Usefulness of Critics}


Participants primarily prioritized critics related to collision risks, joint speed, and space usage (Figure~\ref{fig:combined_results}(right)). Collision detection was ranked higher due to clear safety implications in human-robot interactions. As one participant emphasized, \textit{P8: ``I think the biggest thing I focused on was collisions, because that was the most notable issue, and that [critic description] was just saying, like, hey, make sure you're looking at the collisions.''} 
However, the relevance of certain critics, such as joint speed, was perceived differently among participants. For instance, one participant found joint speed less relevant, assuming the robot would typically operate behind protective barriers, \textit{P12: ``I didn't got bothered by joint speed.''} Overall, the descriptive feedback provided by critics was considered helpful (P11, P19), especially for participants with limited programming experience. One participant noted how it clarified common issues and suggested actionable improvements, \textit{P11: ``I think it will be more difficult because maybe I didn't know how to make sure it correctly.''}. 

Participants appreciated the ease of applying automatic fixes provided by the critics through the buttons, \textit{P18: ``Yeah, the checkers were easy. You can just turn them on or off with a click of a button.''} This ease of use reduced barriers to improving programs, even encouraging some participants to trust automatic solutions without detailed scrutiny, \textit{P19 ``I didn't really analyze it or take it in. Like, I was just being kind of lazy and just thinking like, well, if it fixed the code, it'll work, and if it didn't, I'll click these buttons and try and fix them again.''} Furthermore, some participants noted that reviewing automated fixes from critics helped them gain insights into program improvement and robot control \textit{P11: ``I think maybe I have a little idea that I can change their number or something else to change the move of the robot arm.''}

\paragraph{Tension Between Automation and Control}
Despite the convenience of automatic fixes, participants (P11, P14, P16) expressed a strong preference for greater control over the degree of correction, and often preferred manual text-based input for specific or detailed modifications. They saw manual input as offering precision and flexibility that automatic fixes sometimes lacked, \textit{P11: ``Maybe writing is more easier for me. And because I use this, maybe it will make the robot move quickly or slowly, but if I write... I can do what I want to do.''} 
Occasionally, automatic fixes generated by critics were perceived as overly conservative, leading to user frustration. Some automatic solutions prioritized strict safety constraints at the expense of task completion, leading to confusion or unintended robot behavior, \textit{P16: ``It was more like being stuck trying not to hit the object, but your logical sense is to move above. So somehow the fix was too low level of, it doesn't hit.''} 
Our results suggest that while expert-informed critics significantly enhanced program quality, users' preferences varied considerably, highlighting the importance of balancing automated solutions with user control and clarity in the verification process.

\paragraph{Critic Engagement Patterns}
Critics were sometimes considered secondary to completing the tasks. One participant prioritized initial task completion and saw critics primarily as a tool for subsequent fine-tuning rather than immediate verification, \textit{P14: ``I didn't really use it [critic] too much, because I was trying to get the box in first, and then I thought the end was like, sort of fine tuning, putting restrictions after doing whatever to complete the task, and then just trying to narrow it down.''}

\section{Discussion}


Our findings demonstrate that expert-informed critics substantially improved the reliability of LLM-based robot programming. 
The validation experiment showed that prompt-only embedded critics are insufficient. While embedded in LLM prompts, critics often missed safety-critical violations. External critics, grounded in execution traces, were able to detect and repair these errors, leading to consistently higher program quality. This suggests that critics must operate beyond symbolic code analysis, at the motion level where physical consequences can be directly evaluated. At the same time, persistent challenges such as unresolved pinch points highlight that critics must be complemented with richer robot APIs and better handling of embodied constraints.  

The user study showed that external critics improve program reliability without degrading usability, satisfaction, or workload. Critics were especially helpful for novice users by surfacing safety issues and suggesting corrections. However, participants cautioned against over-reliance on automated fixes, preferring to retain control through manual refinement and selectively engaging with critics they thought most relevant. This selective use suggests that critics serve not only as verification tools but also as guidance for developing safety awareness in robot programming.

\subsection{Design Implications for Integrating Critics}
Together, our results point to three design implications for integrating critics into end-user robot programming systems.

\paragraph{External Verifiers Across Levels of Abstraction}
Our results highlight the capability of external critics to detect safety violations that LLMs, even when prompted with explicit rules, could not reliably identify. This suggests that \textbf{critics should not be confined to static code inspection but instead operate across different layers of abstraction}. Future work should explore designing external critics that span these levels: at the planning level, critics can check task logic, sequencing, and goal satisfaction; at the motion level (as demonstrated by \oursys), they can verify execution traces for collisions and unsafe speeds; and at the runtime level, they can verify code executability. Designing verifiers that span these layers ensures both semantic correctness and embodied safety. Also, while some terminology in the current system remains technical, feedback from critics should be adaptive to user expertise by offering progressively detailed explanations or configurable presentation modes as users gain experience.


\paragraph{Critics and the Automation-Control Tradeoff}
Participants appreciated expert-informed critics that automatically identified and resolved performance or safety issues. Automatic fix buttons simplified program refinement however, some users found these automated fixes overly conservative, highlighting a tension between automation and user control. This trade-off between automation and control underscores the need for careful calibration of automated fixes. \textbf{Future systems should offer flexible levels of automation that adapt to user preferences and experience}. 
Adjustable automation settings can let users balance auto-correction, guided edits, and manual refinement, while transparent explanations of critic-generated fixes help users selectively adopt suggestions. Supporting user-customizable feedback further maintains user control, as demonstrated by prior work that allows users to modulate strictness of constraints in LLM-generated plans \cite{10.1145/3706598.3714113}.

\paragraph{Expanding Skill Libraries for Effective Critic Integration}
The effectiveness of iterative refinement is constrained by the expressiveness of the available APIs. While critics can flag unsafe or suboptimal behaviors, the LLM's ability to respond is limited to the existing skill set. For example, to address a joint-speed violation, the model repeatedly adjusted the parameter of \texttt{reduce\_speed} until no warnings remained, rather than reasoning about alternative motion strategies. This highlights a gap: critics alone cannot resolve embodied issues without sufficient action primitives. \textbf{Future work should focus on expanding robot skill libraries and exposing richer APIs that critics can leverage}. By providing more expressive motion primitives and domain-specific functions, critics can move beyond flagging errors to guiding meaningful fixes, thereby bridging detection with actionable resolution.

\subsection{Limitations and Future Work} \label{sec:limitations}
Our system has several limitations that point to potential areas for improvement. First, the simulation environment was hard-coded, with static environmental information provided to the LLM. Future work could incorporate real-time, dynamic environments that better reflect real-world conditions and synchronization between simulation and sensing.
Second, robot API skills for grasping are hard-coded to simple grip actions, which underscores a reliance on predefined object-centric methods that do not account for grasped geometries. 
Future work should explore richer manipulation representations, including critics that reason about object meshes and geometries, as well as the integration Vision-Language Models (VLMs) and computer vision to support more flexible object manipulation. The critic library could further be extended with perception-aware critics that verify scene consistency and object-state alignment between sensed and simulated environments.
Third, during deployment, issues with the robot's gripper prevented participants from fully observing task execution, even though overall robot movements were evident. Future work should explore the complete pipeline from conceptualization to deployment.
Third, we did not conduct cross-validation of the expert-informed critics with robotic experts. Future work could address this limitation by evaluating critic outputs against expert judgments.
Finally, as an initial exploratory study, our current findings are based on a sample size of 18 participants. While indicative, future work should validate the efficacy of the critic-based integration approach in larger and more diverse populations.

\section{Conclusion}
In this paper, we introduced \oursys, an approach that augments LLM-based robot programming with expert-informed critics, automated fixes, and execution-grounded feedback.
Our evaluation shows that external critics outperform prompt-only verification in detecting safety-critical errors, while a user study reveals that critics improve program quality but are engaged in diverse ways, highlighting tensions between automation and user control. We provide design insights for integrating critics into LLM-based robot programming.

\begin{acks}
This work was supported in part by the National Science Foundation awards 1925043 and 2330040.
\end{acks}

\balance
\bibliographystyle{ACM-Reference-Format}
\bibliography{references}










\end{document}